# SETEASY

*A Multi-Modal Classroom Engagement Assessment and Seating Optimization Framework*

ZHIHAO XIE, HONGYE YANG
*Wuxi Taihu University*
*ghost_aku@163.com*
*Beijing Institute of Architectural Design Co., Ltd*
*yanghongye@biad.com.cn , hyang783@gatech.edu*

AND

SHIEN LIU
*Wuxi Taihu University*
*212017075@wxu.edu.cn*

**Abstract.** SetEasy optimizes classroom engagement in fixed seating grids. It fuses multimodal sensing (wristband physiology, 4K video, environmental data) and trains a v-Gage model grounded in a revised ISEQ. Each week, two-week engagement forecasts are mapped to a student–seat utility matrix, and CP-SAT generates seating plans under visual-access and social-dynamics constraints. In a four-week deployment (23 students, 331 classes), v-Gage converged across affective, behavioral, cognitive, and overall dimensions, cutting RMSE from 0.75 to 0.53. Optimization raised mean engagement from 0.30 to 0.70, with over two-thirds of seats reaching high engagement and back-row low-activity patterns markedly reduced. These results show that, without hardware changes, interpretable, data-driven seating strategies can substantially enhance engagement. The multimodal “assessment + optimization” paradigm offers a transferable, sustainable path to culturally responsive, differentiated spatial design amid global homogenization.

***Keywords:*** *Machine Learning, Educational Space Optimization, Computational Design, Multimodal Sensing, CP-SAT Optimization*

## 1. Introduction

Under the wave of global educational standardization, schools worldwide have largely retained the “grid-based” seating arrangements originating from Western educational systems. This practice frequently results in mismatches between classroom furniture dimensions and students’ body types or cultural needs(Castellucci, Arezes and Viviani, 2010).

In recent years, primary and secondary education reforms have called for a shift in learning environments from unidirectional instruction to learner-centered models that emphasize individual differences and foster active collaboration(Barrett et al., 2015, Yang and Guttmann-Flury, 2026). Extensive research indicates that classroom seating arrangements and spatial design have a significant impact on students’ learning outcomes and classroom engagement(Mäkelä and Leinonen, 2021). For example, Barrett et al. found that simply optimizing seating layout can yield up to a 16% improvement in learning efficiency. However, limited by school infrastructure and funding, most schools still rely on fixed desk layouts, making it a practical challenge to optimally assign “who sits where” within the existing grid(Peponis et al., 2007, Yao et al., 2024, Yang, Liu and Xie, 2026).

Currently, seat assignments in classrooms primarily rely on teachers’ experience—either by making manual adjustments based on guidelines and shortest-path methods(Meller and Gau, 1996), or by iteratively experimenting with adjacency matrices in CAD software(Liggett, 2000). When balancing multiple objectives such as capacity, visibility, and social dynamics, this process is time-consuming and rarely achieves a globally optimal outcome. Fixed seating layouts have also been shown to inhibit collaboration and behavioral engagement(Kariippanon et al., 2019).

In response to these limitations, research in computationally intelligent seat assignment has advanced rapidly. Genetic algorithms are capable of searching thousands of combinations at once, significantly improving satisfaction(Shin-Ike and Iima, 2016); Merrell et al. mapped interior design principles into probability density functions to interactively recommend seating positions(Peponis et al., 2007); deep reinforcement learning has modeled seat swaps as a Markov process, converging on multi-objective optima(Kakooee and Dillenburger, 2024); and the EDU-AI framework applies attention heatmaps in closed-loop seating optimization(Kariippanon et al., 2019). However, most existing methods depend on single behavioral factors or static preferences and lack real-time integration of multimodal data—physiological, behavioral, and environmental.

Meanwhile, educational scholars have long criticized “one-size-fits-all” classroom models for undermining students’ cultural identity and learning motivation(Ladson-Billings, 2014). Culturally responsive teaching stresses a deep understanding of learners' sociocultural backgrounds(Gay, 2002), and self-determination theory emphasizes that environments must satisfy students' needs for autonomy, competence, and relatedness(Pekrun and Linnenbrink-Garcia, 2012). The OECD likewise advocates for locally data-driven design to break the homogeneity of seating arrangements(OECD, 2013). Yet, teachers working in classrooms with fixed desks and chairs still lack quantitative tools for implementing differentiated seat assignment.

To bridge this gap, we present SetEasy—an intelligent framework for classroom seating optimization. The system first integrates physiological and behavioral data from students and teachers, as well as environmental information, and incorporates questionnaire results to construct an engagement prediction model. Combined with seat calibration data, these inputs are used to generate a student-seat utility matrix, enabling the generation of optimal seating plans through a weekly closed-loop iteration. In real-world deployment, after seven cycles, overall classroom engagement increased by approximately 45%, validating that data-driven seat assignment can significantly enhance educational outcomes in traditional fixed seating environments. Furthermore, SetEasy provides a scalable and replicable technical pathway for implementing localized and culturally responsive classroom practices.

## 2. Methodology

SetEasy is an intelligent, multimodal perception-based framework for optimizing seating arrangements within fixed classroom grids. The system first collects data from wristband physiological sensors, 4K behavioral video, and environmental monitors, and integrates these with ISEQ questionnaire responses to build the v-Gage engagement prediction model. Engagement predictions from the previous two weeks are fused with seat calibration data to generate a student-seat utility matrix. Finally, CP-SAT integer programming is used to maximize utility under teacher-defined constraints, such as vision and social dynamics, outputting the optimal seating plan in a weekly closed-loop iteration (see Figure 1).

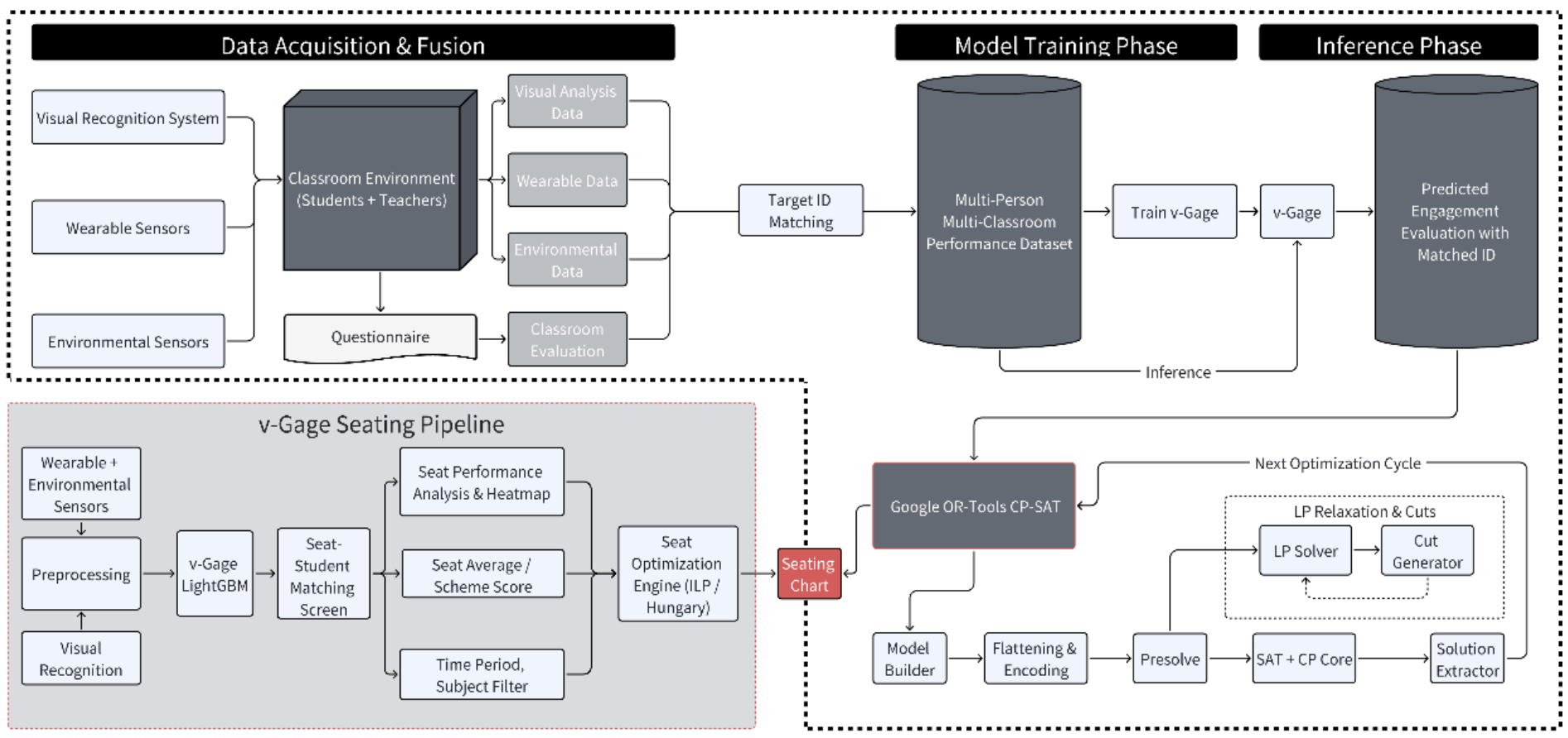

*Figure 1.* Classroom Engagement & Seating Flow.

## 2.1 DATA COLLECTION

Data were collected in June 2025 from a standard classroom at a secondary school in a medium-sized city in southern China. Over a four-week period, 23 first-year high school students (13 female, 10 male) and 6 teachers participated as study subjects, resulting in 331 valid classroom session datasets.

### *2.1.1 Physiological and Behavioral Data Collection*

Three types of sensors were deployed for data collection (see Figure 2). First, each student wore an Empatica E4 wristband to continuously capture multi-channel physiological signals during class, including electrodermal activity (EDA) at 4 Hz, blood volume pulse (BVP) at 64 Hz (used to derive heart rate variability, HRV, metrics), 3-axis acceleration (ACC) at 32 Hz, and peripheral skin temperature (ST) at 4 Hz(Gao et al., 2020).

Second, to obtain fine-grained behavioral data, the system followed the approaches of the StuArt model(Zhou et al., 2023) and the open-source Student Classroom Behavior dataset(Yang, Wang and Wang, 2023), performing real-time recognition of five key classroom behaviors—hand-raising, standing, writing/reading, dozing, and mobile phone use—at a frequency of once per second. A 4K wide-angle camera was mounted at the top front of each classroom to continuously record and analyze classroom dynamics at one frame per second.

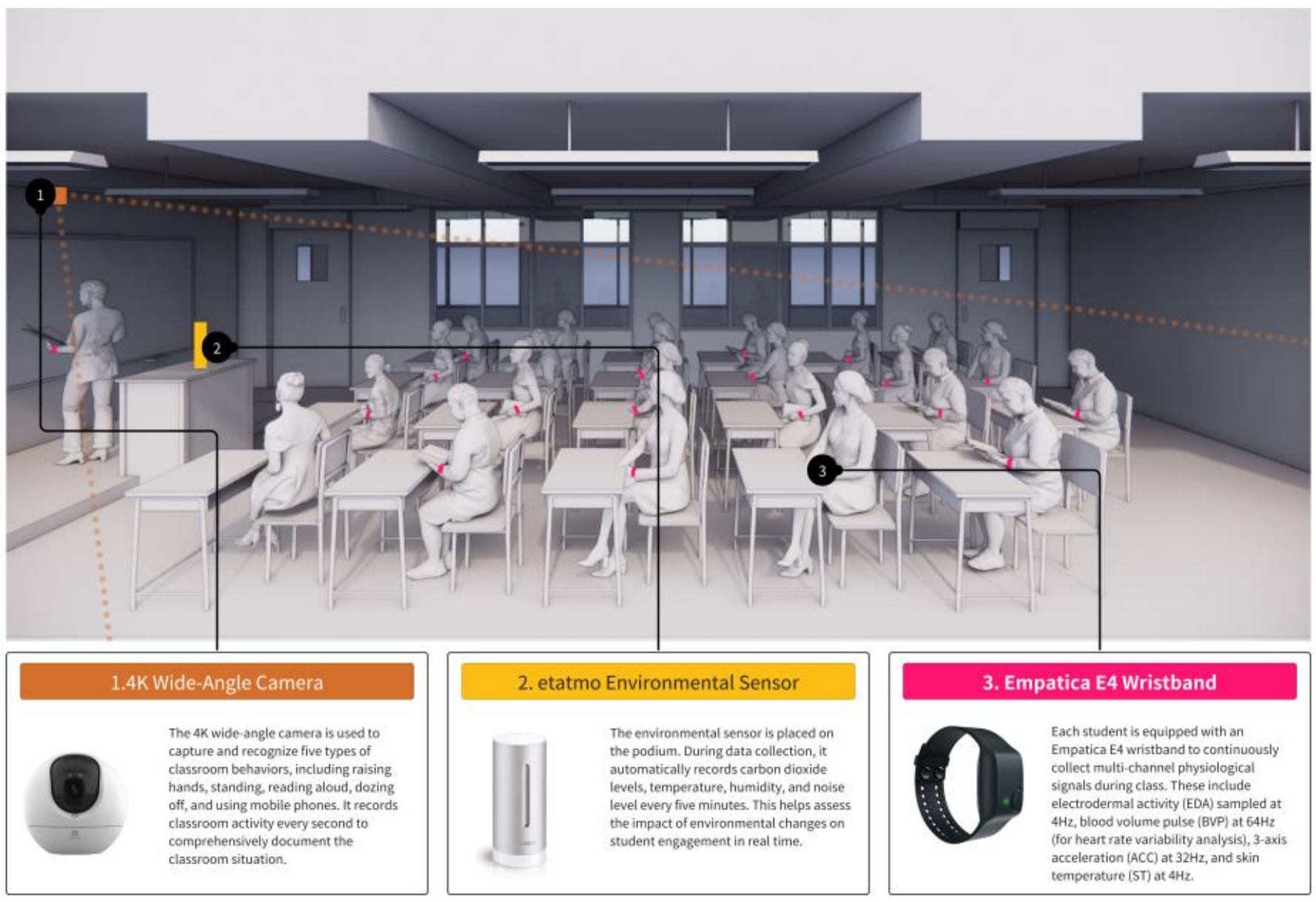

*Figure 2.* Diagram of Physical Deployment in Classroom.

Additionally, each classroom was equipped with a Netatmo environmental sensor, automatically recording indoor $CO_2$ concentration, temperature, humidity, and noise level every five minutes to provide real-time assessment of environmental impacts on students. The overall data sensing timeline is shown in Figure 3.

We implement multiple privacy safeguards: (1) guardians and students provide written informed consent; (2) video and wearable data are processed exclusively via on-premises, offline inference on school computers—raw data are never written to disk and are immediately deleted, with only de-identified weekly features and seat scores retained; (3) students participate under randomly assigned ten-digit IDs, reducing re-identification risk.2.2

### *2.1.2 Engagement Annotation and Ground Truth Construction*

In this study, "classroom engagement" is defined as a composite of students' focus and willingness to participate in classroom activities across three dimensions:

(1) Cognitive engagement: the depth of students' understanding and critical thinking about classroom content;

(2) Affective engagement: the level of interest and enjoyment students experience during lessons;

(3) Behavioral engagement: students' attentiveness and active participation in class.

Although multimodal sensor data provide objective measurements of students' behaviors and physiological states, subjective self-report measures remain crucial for supervising the predictive model. To ensure reliable ground truth, we adopted the validated In-Class Student Engagement Questionnaire (ISEQ)(Fuller et al., 2018).

Consistent with previous research using wearables in classroom settings, which adapts survey items for different age groups(Di Lascio et al., 2017), we simplified the original ISEQ—designed for college students—to better suit the cognitive level of high school students.

Additionally, based on recommendations from Moore and Lippman regarding positive development indicators, we replaced the ISEQ item "the activities really helped my learning of this topic" with "I asked myself questions to make sure I understood the class content," thereby minimizing scoring bias in courses with limited in-class activities.

Table 1 presents the questionnaire items used to assess multidimensional engagement: items 1, 3, and 5 measure behavioral, affective, and cognitive engagement, respectively, while items 2 and 4 reflect disengagement in behavior and emotion(Skinner, Kindermann and Furrer, 2009, Fuller et al., 2018).

All items were rated on a five-point Likert scale (－2 to +2), and students completed the questionnaire immediately after each class to minimize recall bias. The questionnaire data collection process is illustrated in Figure 3.

TABLE 1. Self-report items for measuring in-class engagement in online survey.

| Questions (please describe your engagement in the last class) | Subscales |
|---|---|
| 1. I paid attention in class. | Behavioural |
| 2. I pretended to participate in class but actually not. | Behavioural (-) |
| 3. I enjoyed learning new things in class. | Emotional |
| 4. I felt discouraged when we worked on something. | Emotional (-) |
| 5. I asked myself questions to make sure I understood the class content. | Cognitive |

Note: (-) means the reversed score.

During data processing, all reverse-coded items were first adjusted by inverting their scores, after which the raw scores were linearly transformed to a 1–5 scale. The arithmetic mean of the cognitive, affective, and behavioral scores was then calculated to yield the overall engagement ground truth for each class session.

## 2.2 DATA PREPROCESSING

To accurately determine the start and end times of classroom sessions, we applied Information-Gain based Temporal Segmentation (IGTS)(Sadri, Ren and Salim, 2017). This method automatically differentiates between instructional and break periods based on student activity patterns, ensuring that all subsequent analyses are strictly confined to genuine classroom teaching intervals.

To address prevalent motion artifacts and environmental noise in the physiological data, we implemented a multi-stage cleaning process: EDA signals were first smoothed using a 5-second median filter to suppress transient noise, then decomposed into tonic and phasic components via the cvxEDA algorithm. For BVP data, linear interpolation was employed to repair intervals with suboptimal heart rate quality.

Following the removal of flat segments and artifact contamination, we validated a total of 331 high-quality, usable class sessions, which served as the foundational dataset for further analysis.

### 2.3 FEATURE-ENGAGEMENT ENGINEERING

The central goal of this study is to construct effective features for precise classroom engagement prediction. Building on related work such as n-Gage(Gao et al., 2020, Zhou et al., 2023), we replicated and enhanced the multidimensional engagement prediction pipeline to fit our research context, resulting in the new v-Gage system module.

From physiological signals, we extracted 36 features, including the amplitude and number of peaks in EDA signals, proportion of arousal, and HRV indices in both time (SDNN, RMSSD, pNN50) and frequency (LF/HF ratio, etc.) domains(Shaffer and Ginsberg, 2017). For behavioral patterns, we calculated the mean and variance of motion intensity from accelerometer data, and constructed synchrony features with both teachers and peers using Dynamic Time Warping (DTW) and Pearson correlation coefficients (8 features total).

From classroom environmental data, we extracted 8 features—mean and peak values of $CO_2$, temperature, humidity, and noise during lessons—to quantify environmental impacts on students(De Dear and Brager, 2002).

The visual behavior recognition stream further enriched the system with direct classroom behavioral features: for each class session, we detected the frequency and duration of the five most common behaviors, hand-raising-to-teacher-response latency, and cumulative dozing time, resulting in 10 behavioral features per session. Altogether, this yielded a comprehensive feature set with 62 dimensions.

We employed a LightGBM (Luxburg et al., 2018)regression model to model the relationship between the feature set and students' classroom engagement. The LightGBM-based v-Gage automatically predicts three-dimensional engagement, offering a rapid, efficient, and quantifiable alternative to traditional subjective scoring. Model training followed a nested cross-validation scheme: the inner loop used three-fold cross-validation for hyperparameter tuning and feature selection, while the outer loop applied five-fold cross-validation (grouped by student ID) to ensure generalizability and prevent information leakage.

### 2.4 CONSTRUCTION OF THE STUDENT–SEAT UTILITY MATRIX

To map multimodal predictions of individual classroom engagement to actual teaching space, we constructed a student – seat utility matrix that quantitatively evaluates the potential effects of every possible seating arrangement on classroom outcomes. First, we performed spatial calibration of the classroom using a single overhead 4K camera and ARUCO marker technology(Garrido-Jurado et al., 2014), enabling one-time calibration of the classroom layout and seat grid and assigning spatial coordinates and unique indices to each seat. During each class session, the visual model ran a real-time head tracking algorithm and employed the SORT multi-object tracking method(Bewley et al., 2016) to accurately map student IDs in every frame to their respective seats, allowing the system to automatically generate and store student – seat assignment sequences for each session.

Based on these spatial calibration and seat-matching results, we aggregated the predicted engagement of each student for every seat over the preceding two weeks. Specifically, we calculated the average predicted engagement ($\bar{E}_{i,j}$) for each student – seat pair to reflect long-term trends in classroom performance, and computed the standard deviation ($\sigma_j$) for each seat to quantify the stability of engagement across different students. We then combined these metrics to define a comprehensive utility score:

$$U_{i,j} = 0.8\bar{E}_{i,j} + 0.2(1/\sigma_j) \quad (1)$$

The weighting coefficients of 0.8 and 0.2 were selected based on sensitivity analysis and cross-validation, emphasizing the long-term performance of individual seats while appropriately considering the consistency and reliability of each seat's engagement level. This utility matrix serves as a crucial input for the subsequent seat optimization model, quantifying the influence of classroom spatial configuration on overall classroom engagement.

### 2.5 SEAT OPTIMIZATION SOLVER

Building on the established seat utility matrix, we rigorously modeled the seat assignment as an integer linear programming problem and solved it using the Google OR-Tools CP-SAT solver(Martinelli Tabajara and Y. Vardi, 2019) for efficient optimization. The decision variable $x_{i,j} \in \{0, 1\}$ is set to 1 if student i is assigned to seat j, and 0 otherwise. The objective function is to maximize the total weighted utility across all students:

$$\max \sum_{i,j} w_i \; U_{i,j} \, x_{i,j} \tag{2}$$

Here, the weight $w_i$ is customizable by the teacher to address specific pedagogical needs — such as assigning higher weights to students with learning difficulties or attention deficits—enabling proactive support for these students through the optimization process.

The optimization is subject to the following constraints:

(1) Each student must be assigned to exactly one seat (student-seat uniqueness);

(2) Each seat may be assigned to at most one student (seat-student exclusivity);

(3) Vision and height constraints: students with impaired vision or shorter stature are prioritized for front-row seats with unobstructed views;

(4) Social and collaboration constraints: avoid seating students who are prone to mutual distraction or negative interactions next to each other, while promoting proximity among students who collaborate effectively;

(5) Accessibility and special needs: ensure that students with disabilities or special requirements are seated in accessible locations;

(6) Teacher-specified reserved or restricted seating zones, providing teachers with necessary intervention capabilities.

## 2.6 MODEL DEPLOYMENT

To evaluate the usability of the system in authentic classroom settings, all algorithm modules were deployed on the standard classroom computer used by teachers, with automatic data synchronization and analysis handled over the local area network(Xie et al., 2019, Liu et al., 2022).

The classroom was equipped with a standard teaching computer (Intel Core i5, 16 GB RAM, NVIDIA RTX 3060, Ubuntu 20.04), serving as the central computation and storage node. All wearable physiological sensors, environmental monitors, and classroom cameras were connected to the same Wi-Fi LAN and configured to automatically upload collected data to a designated directory on this computer after each class, ensuring comprehensive and centralized data management.

A unified Python 3.8 virtual environment was created on the deployment machine, with required dependencies — including PyTorch, OpenCV, LightGBM, Scikit-learn, and OR-Tools — installed sequentially. The final model weights and configuration files for the behavior recognition, engagement prediction, and CP-SAT seat optimization modules were synchronized locally. Each configuration file specified the data input directory, model path, and output directory to guarantee that inference results matched those obtained during the laboratory phase.

Once configured, the system operates automatically according to the following workflow:

(1) After each class, all devices upload data;

(2) The classroom computer sequentially performs behavior recognition, engagement prediction, and seat optimization;

(3) Results are saved in the local database and reports are generated.

### 2.7 PERIODIC CLOSED-LOOP OPTIMIZATION

SetEasy is designed not as a one-time static deployment but as a periodic, closed-loop system supporting continuous, dynamic optimization in real classroom settings.

At the end of each week, SetEasy automatically launches a complete data update and optimization workflow. The ETL pipeline aggregates logs from wearable devices, raw data from environmental sensors, and outputs from visual behavior recognition into a unified time-series database. Data preprocessing — including normalization, missing value imputation, and feature selection (such as moving average smoothing and PCA-based dimensionality reduction)—is applied within the data warehouse(Moreno-Marcos et al., 2019, Choi, Yang and Chung, 2021). The engagement prediction model is incrementally updated using gradient descent, employing sliding window cross-validation on data from the past seven days, with optional small-scale hyperparameter tuning (e.g., learning rate adjustment, regularization tuning) to maintain model stability.

The updated model generates utility scores for each student at each seat, creating a new sparse matrix as input to the Google OR-Tools CP-SAT solver. The solver aims to maximize the overall mean engagement, incorporating linear constraints (such as group distance, special-needs seat masks, and front–back row ratios) and binary decision variables, and conducts time-limited heuristic search to output optimal or near-optimal seating solutions.

The system's frontend component visualizes the results as a heatmap covering the entire classroom, accompanied by key decision prompts. Teachers can confirm or manually adjust the recommended arrangement, with all changes recorded and integrated into the next cycle's dataset—thus closing the loop. Through incremental model updates, matrix reconstruction, CP-SAT solving, and visual analytics, SetEasy delivers a robust and reproducible workflow for ongoing dynamic seat optimization:

(1) After each cycle, the engagement prediction model is automatically updated using the previous week's classroom data;

(2) The student－seat utility matrix for the next cycle is recalculated according to the updated model;

(3) The CP-SAT optimization engine generates seating recommendations and corresponding heatmaps, providing intuitive explanations for each suggestion;

(4) Teachers can confirm the layout or interactively adjust the recommendations provided by the system.

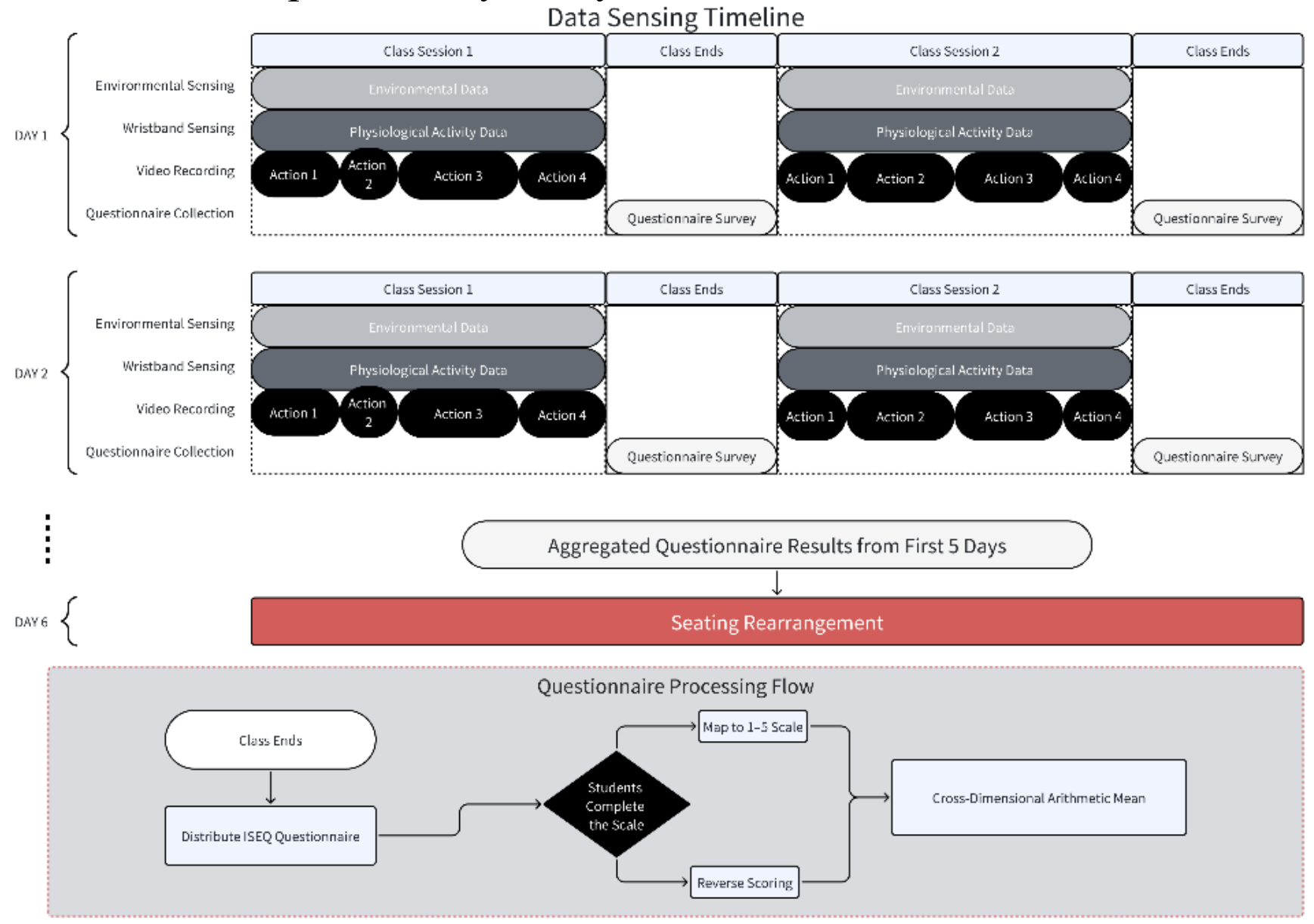

*Figure 3.* Data & Questionnaire Flow.

## 3. Results

### 3.1 DATA COLLECTION OUTCOMES

In a representative lesson—a 45-minute mathematics class during week 1, session 1—the SetEasy system automatically captured and de-duplicated five categories of classroom behaviors, retaining seat numbers (RxCy): hand-raising, standing, yawning, smiling, and brief episodes of dozing (see Table 2). Across the session, the 23 students collectively exhibited 230 hand-raises (an average of approximately 10 per student), 45 instances of standing (about 2 per student), 60 yawns (about 2.6 per student), 180 smiles (about 7.8 per student), and 8 dozing events (about 0.3 per student). Hand-raising activity peaked between the 10th and 30th minutes, coinciding with periods of teacher explanation and class interaction; standing events were primarily associated with roll call or demonstration scenarios, averaging 5.5 seconds in duration;

yawning and dozing occurred mainly in the second half of the class, suggesting a decline in student attention over time.

Further spatial analysis revealed systematic differences in behavioral frequency by seat location (see Table 2). Students seated in the front row (e.g., R1C2, R1C3) raised their hands 11 – 12 times on average, while those in the back row (R5C3, R4C1) did so only 8 – 9 times. In contrast, yawning and dozing were more frequent in the back rows; for example, R3C4 recorded 3 yawns and 1 dozing event, whereas R1C1 in the front row recorded no sleep behavior. Smiling also showed a front-row-high, back-row-low pattern: R1C2 and R2C3 registered 9 – 10 instances, while R4C3 in the back row registered only 6. These results objectively demonstrate that, within a fixed seat grid, spatial distance significantly influences students' active participation and fatigue-related behaviors.

TABLE 2. Seat-ID-Behavior Statistics.

| Seat (RxCy) | Stu ID | Raising Hands | Standing | Yawning | Smiling | Sleeping |
|---|---|---|---|---|---|---|
| R1C1 | S01 | 9 | 2 | 1 | 8 | 0 |
| R1C2 | S02 | 12 | 2 | 0 | 10 | 0 |
| R1C3 | S03 | 11 | 1 | 1 | 9 | 0 |
| R1C4 | S04 | 10 | 3 | 2 | 7 | 1 |
| …… | …… | ...... | …… | ...... | ...... | ...... |
| R6C3 | S23 | 9 | 1 | 1 | 9 | 0 |

3.2 V-GAGE MODEL EVALUATION

As shown in Figure 4, the MAE and RMSE curves for v-Gage in emotional, behavioral, cognitive, and overall engagement dimensions exhibit a steady, monotonic decline, with no oscillations or inflection points, indicating stable convergence and controlled overfitting risk during training. Compared to n-Gage—which utilizes only physiological and environmental features—the addition of behavioral features in v-Gage yields consistently lower prediction errors across all dimensions. The most notable improvement is in cognitive engagement: after 80 training epochs, v-Gage achieves an RMSE of approximately 1.00, while n-Gage remains at 1.11. The overall engagement RMSE also drops from 0.75 to around 0.53, demonstrating superior generalization. This improvement is mainly attributable to the incorporation of behavioral features such as acceleration intensity and group synchrony, which are highly correlated with cognitive workload and thus effectively enhance the expressive power of physiological and environmental signals.

### 3.3 SEAT HEATMAP EVALUATION

A comparison of seat heatmaps before and after optimization (Figure 4) clearly demonstrates that reassigning seats substantially increased overall class engagement while reducing spatial variability. Prior to optimization, most seats had engagement scores in the 0.20 – 0.40 range, with a global average near 0.30 and the highest value barely reaching 0.60. The third row recorded the lowest average (≈ 0.35) and included near-zero "non-participation" seats, reflecting a significant proportion of low-activity locations.

After optimization, every seat except row 4, column 2 exceeded 0.60, lifting the overall class average to approximately 0.70. More than two-thirds of seats registered scores above 0.80, with continuous "high-engagement bands" appearing in rows 1, 3, and 6, indicating clear convergence in engagement distribution. The change in average engagement (0.30 → 0.70) represents an increase of over 130%, and only a single seat remained at an extremely low value, showing that "low-participation islands" were nearly eradicated.

Overall, data-driven seat optimization shifted engagement from "low and dispersed" to "high and concentrated," providing teachers with clear targets for further improvement and validating the effectiveness of seating optimization in enhancing classroom engagement.

## 4. Discussion

### 4.1 SUMMARY OF RESULTS

Applied in mainland Chinese secondary classrooms—across both Chinese- and English-language lessons—SetEasy received positive feedback: The results demonstrate that SetEasy uncovers significant temporal and spatial stratification in classroom engagement: student participation exhibits a "rise-then-fall" pattern over the course of a lesson, and positive behaviors become sparser and signs of fatigue more evident with increased distance from the teacher. The v-Gage model achieves stable, smooth convergence in multidimensional engagement prediction; the integration of behavioral features substantially enhances the identification of deep cognitive states, validating the complementarity between physiological–environmental signals and overt behaviors. With this detailed engagement profile, integer programming reallocates high-potential students to key sightlines, elevating overall classroom engagement and leading to a more balanced distribution: previous patterns of "low participation in the back row" are effectively alleviated, with high-engagement areas clustering into continuous bands and low-activity "islands" significantly reduced.

In summary, whether in Chinese or English classes, SetEasy first diagnoses seat-induced disparities in participation and then, through data-driven seat assignment, consolidates engagement from a scattered state to one that is concentrated and sustainable—demonstrating the synergistic effect of multimodal sensing and optimization strategies in reconstructing classroom motivation dynamics.

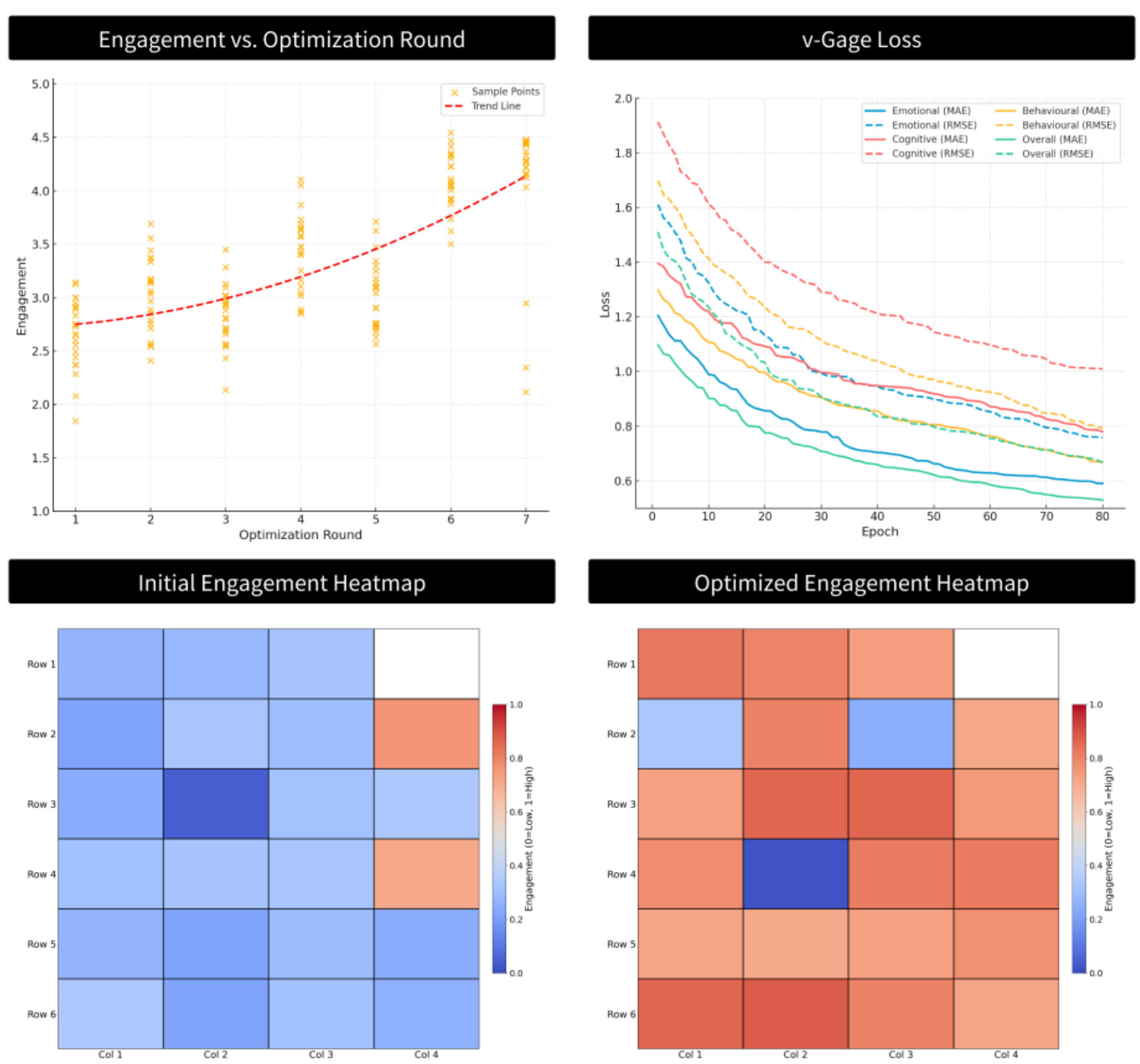

*Figure 4.* Visualization of Model Training and Seating Optimization Outcomes.

## 4.2 LIMITATIONS

SetEasy focuses on lecture classrooms because the seating and visibility gaps it corrects rarely occur in seminars or labs. In this context, this study has four primary limitations. First, the model was trained mainly for standard lecture classrooms and has not yet been validated in diverse instructional contexts such as inquiry-based or discussion-oriented classes, nor has its adaptability to different teaching styles been established. Future work should broaden the range of classroom types and teacher samples to enhance generalizability. Second, the engagement ground truth relies in part on student self-assessment, which may be affected by fatigue and social desirability bias; future studies should include multi-source evaluations from peers and teachers, along with more objective metrics, to mitigate subjective error. Third, data collection was limited to a small number of campuses and grade levels; thus, the model's robustness in other locations, educational stages, or non-standard classroom layouts remains to be demonstrated. Systematic evaluation of cross-context transferability, enabled by cross-school collaboration and data sharing, will be necessary for further refinement. Fourth, data collection required many wristbands and an ambient sensor, so large-scale rollout is costly.

### 4.3 CONCLUSION

SetEasy integrates multimodal sensing, v-Gage engagement prediction, and CP-SAT integer optimization, enabling the calculation of "student–seat" utility within fixed seating grids and providing weekly, class-wide seat assignment strategies that maximize collective benefit. This framework equips teachers with interpretable, quantifiable, and rapidly updatable seating strategies, replacing labor-intensive manual adjustments and enhancing both individualized support and instructional efficiency. Instructors reported increased interaction and reduced management burden with higher student engagement, aligning with our quantitative results. More importantly, this study demonstrates that even when physical layouts cannot be altered, computational design can reshape spatial dynamics and mitigate the suppression of local cultural identity and motivation imposed by globalized, homogenized layouts—laying a data-driven foundation for culturally responsive and differentiated design in educational architecture. This "multimodal assessment–integer optimization" paradigm is equally applicable to other fixed-seat environments, such as meeting rooms, control centers, and medical waiting areas, offering a transferable technical template for smart space management and equitable resource allocation.